\pdfoutput=1

\documentclass[11pt]{article}

\usepackage[final]{coling}

\usepackage{times}
\usepackage{latexsym}

\usepackage[T1]{fontenc}

\usepackage[utf8]{inputenc}

\usepackage{microtype}

\usepackage{inconsolata}

\usepackage{graphicx}

\usepackage{algorithmicx}
\usepackage[ruled]{algorithm2e}
\usepackage{algpseudocode}
\usepackage{bbm}

\title{Enhancing Mathematical Reasoning in LLMs with Background Operators}

\author{Jiajun Chen$^*$ \and Yik-Cheung Tam$^*$\\
Shanghai Frontiers Science Center of Artificial Intelligence and Deep Learning\\
New York University Shanghai\\
\{\href{mailto:jc11815@nyu.edu}{jc11815},\href{mailto:yt2267@nyu.edu}{yt2267}\}@nyu.edu}
  
\usepackage{float}
\usepackage{amsmath} 
\usepackage{booktabs} 
\usepackage{longtable}
\begin{document}
\maketitle
\def\thefootnote{*}\footnotetext{Equal contribution.}\def\thefootnote{\arabic{footnote}}
\begin{abstract}
We propose utilizing background operators for mathematical reasoning in large language models (LLMs). To achieve this, we define a set of fundamental mathematical predicates as the basic building blocks. For each mathematical problem, we develop a Prolog solution that includes problem-specific predicates and intermediate predicates derived from these background operators, ensuring that each solution adheres to the defined operator set. We introduce the MATH-Prolog corpus, which is derived from the counting and probability categories of the MATH corpus. For efficient data augmentation, we apply K-fold cross-validated self-training. This method incrementally generates new Prolog solutions for each fold, incorporating those verified as correct into the training set throughout the model training process. Our experimental results demonstrate that 5-fold cross-validated self-training effectively identifies new, accurate Prolog solutions, achieving an accuracy of 84.6\% on the cross-validated set, and 84.8\% on the test set during fine-tuning the Meta-Llama-3.1-8B-Instruct model. This approach successfully uncovers new solutions with fully computable inference steps for previously unseen problems. Additionally, incorporating the background mathematical predicates into the prompt enhances solution coverage.
\end{abstract}

\section{Introduction}
Recently, large language models have shown success in tackling mathematical problems such as chain of thought~\cite{wei2022chain,deepseek-math}, tree of thought~\cite{yao2023tree}, program of thought~\cite{chen2023program}, and graph of thought~\cite{besta2023graph}. For example, chain of thought has an advantage that each reasoning step is described verbally which is good for human understanding. However, verbal reasoning steps are not computable and therefore hard for machines to verify for flaws. Generating computer programs with Python codes such as Sympy for mathematical reasoning is promising~\cite{gou2023tora}. However, this approach has limitation for logical reasoning since the programming language is procedural in nature. Using Prolog language to describe a mathematical problem would be more powerful. Solving a mathematical problem is converted into searching for a final answer such that problem-specific constraints are satisfied.

\begin{figure}[t]  
  \centering
  \includegraphics[width=\columnwidth]{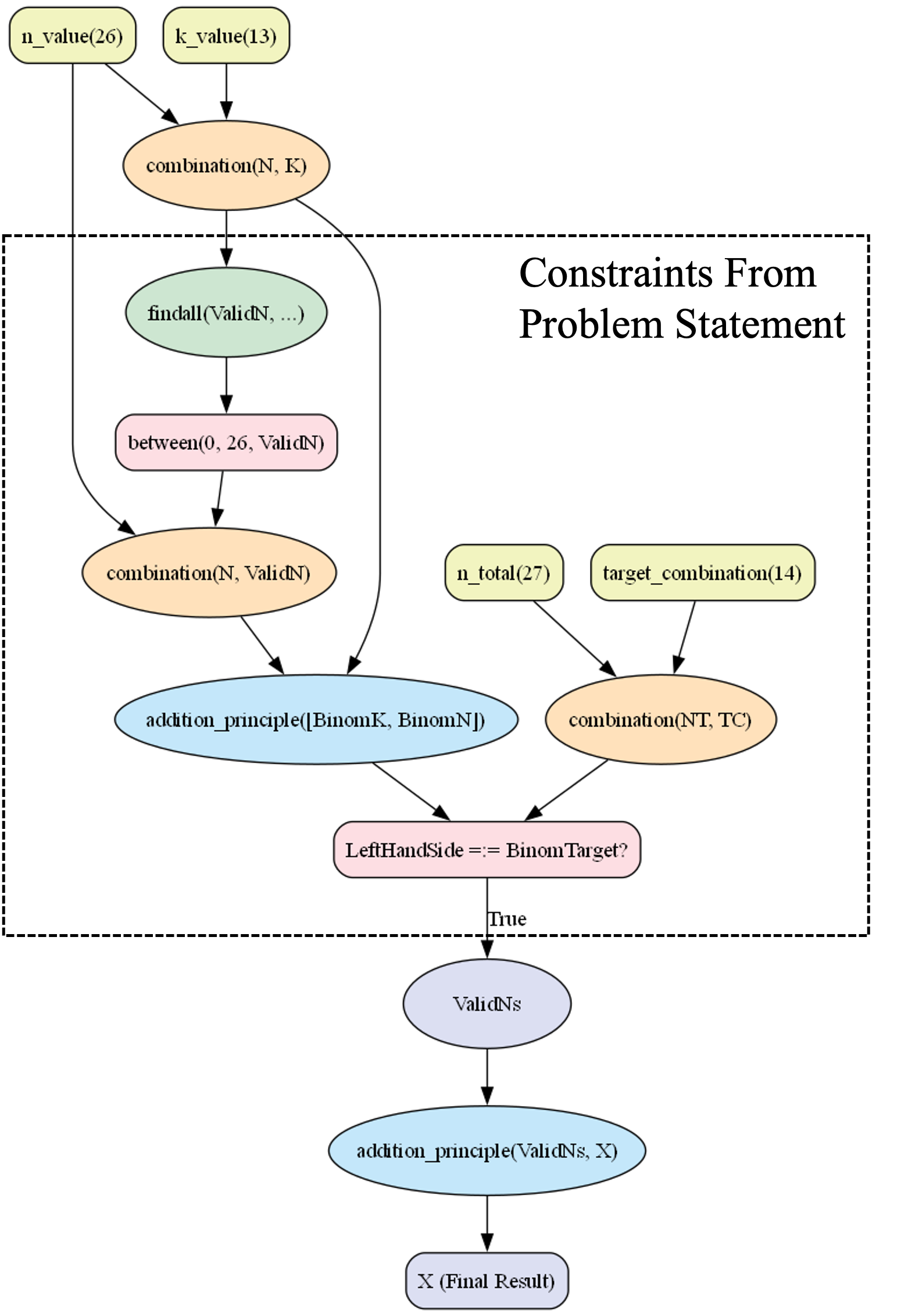}  
  \caption{Computation graph of a Prolog solution for the problem ``What is the sum of all integer values $n$ for which $\binom{26}{13}+\binom{26}{n}=\binom{27}{14}$?''.}
  \label{fig:cg}
\end{figure}
In this paper, we cast solving mathematical reasoning problems via Prolog which is realized as a set of predicates. A search process facilitated by an external interpreter, e.g. Prolog, is performed to find the feasible values of the predicates serving as constraints. Each constraint in Prolog must be satisfied to arrive at a final answer. Many questions in the MATH corpus can be casted as search problems satisfying a set of constraints. For example, for a problem ``If two numbers are randomly chosen without replacement from $\{1, 2, 3, 4, 5\}$, what is the probability their sum is greater than their product?'', the constraints are: 1) The two numbers $A$ and $B$ belongs to the set $\{1, 2, 3, 4, 5\}$; 2) $A+B>A \cdot B$; 3) $A < B$. We want a language model to write Prolog code that declares the constraints and search for a set of feasible solutions $\{(A,B)\}$ to compute the final probability.

To enable a large language model to generate Prolog code, our proposal is to curate MATH-Prolog, a corpus for solving mathematical reasoning problems based on Prolog. Then, we fine-tune a large language model on the MATH-Prolog to learn how to generate Prolog codes. To enforce standardized solutions, we curate a set of pre-defined background mathematical predicates, e.g. $combination(n,k,out)$ and $factorial(n,out)$ as standard predicates so that all Prolog solutions must be constructed based on these background operators. 
As another viewpoint, two predicates $A(X_1, X_2, Y),B(Y, Z)$ takes two input symbols $X_1$ and $X_2$ and outputs $Y$ which is an input to predicate $B(Y, Z)$ resulting in an output $Z$. Therefore, a Prolog solution is an instance of a directed computation graph as illustrated in Figure~\ref{fig:cg}, where the input nodes define the problem-specific predicates inferred from the problem statement, and the intermediate nodes are the background operators which are connected with each other so that the final result $X$ is arrived. A pretrained large language model is finetuned to generate a computation graph to solve a mathematical reasoning problem.

Our research has the following contributions: 1) We curate and open-source the MATH-Prolog dataset focusing on counting and probability to enable large language models to generate declarative Prolog codes to solve competition-level mathematical reasoning problems. 2) We devise a cross-validated self-training algorithm that generates and explores Prolog code solutions during model finetuning on unseen problems. 3) Experimental results show that cross-validated self-training algorithm generates codes that lead to 84.8\% accuracy on the test set.

\section{Corpus Design}\label{sec:Corpus}
We design the MATH-Prolog based on the MATH corpus, a benchmark for competition-level mathematical reasoning problems. We choose counting and probability in MATH for corpus curation consisting of 771 problems on the training set. For each problem in the training set, we curate a Prolog solution which consists of two parts: 1) Problem-specific predicates which are directly inferred from the problem statement; 2) The Solve predicate composing the intermediate predicates as constraints required to solve the given problem. 
In total, we have 54 background operators manually crafted from high-school textbooks, and all Prolog solutions are standardized to this operator set. 
Some background operators such as $palindrome(n)$ are included since some training questions require the palindrome constraint.
With the help of GPT-4 and few-shot prompting technique, we handcraft Prolog solutions of a subset of the problems in the training set, and use GPT-4 to generate Prolog solutions on the rest of training problems following~\cite{yang-etal-2024-arithmetic}. Since the MATH corpus is competition level, many of the GPT-4 generated solutions actually do not evaluate to correct final answers. Therefore, we manually check the generated solutions and make code adjustments so that the evaluated Prolog codes result in correct  reference answers. This costs one manual labor for about one month to curate the high-quality and 100\% accurate MATH-Prolog training corpus for research. Due to limited manual resources, we did not manually curate the 474 problems on the test set. Inspired by~\cite{wang2024dpost}, we employ a self-training algorithm to automatically generate Prolog solutions for the test set.
\subsection{Findall Predicate}
\begin{figure}[htbp]
  \centering
  \includegraphics[width=\columnwidth]{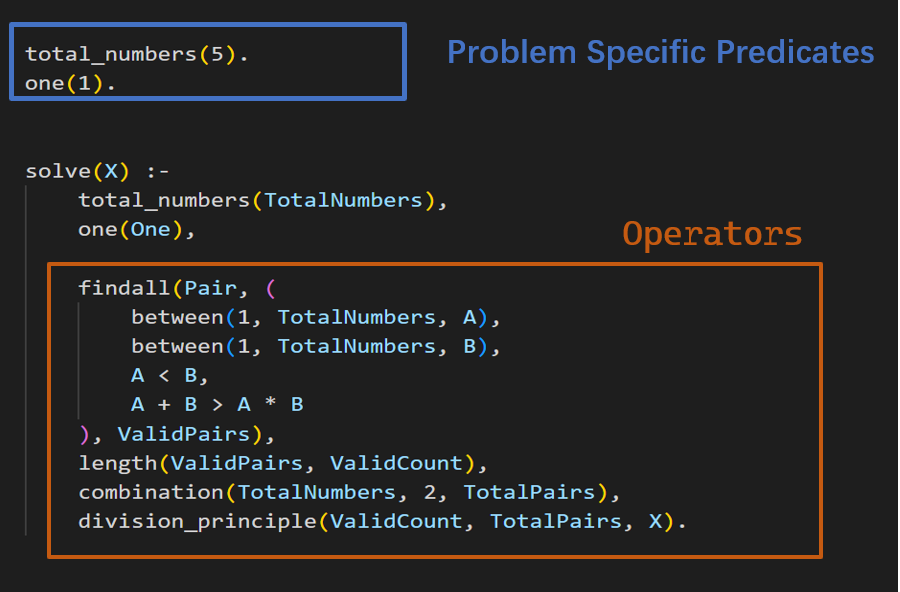}  
  \caption{A Prolog code solution for the problem ``If two numbers are randomly chosen without replacement from $\{1, 2, 3, 4, 5\}$, what is the probability their sum is greater than their product?''.}
  \label{fig:prolog_code_2}
\end{figure}

Given that we have a finite set of problem-specific predicates and a finite set of predicates, the correct solution can be derived systematically by assembling the variables and predicates. From a graphical perspective, these variables and predicates can be viewed as nodes, with directed edges representing the flow of inputs and outputs within the solution. However, exhaustively traversing all possible combinations of variables and predicates is computationally expensive. To address this challenge, we employ the $findall(.)$ predicate to represent the constraints derived from the problem statements, effectively narrowing down the search space. $findall(.)$ is a standard predicate defined in SWI-Prolog.
By leveraging the $findall(.)$ predicate, we can express the relevant constraints in a compact manner.
Figure~\ref{fig:prolog_code_2} gives an example of the usage of $findall(.)$ predicate that returns a list of valid pairs $(A,B)$ satisfying the constraints required by the problem.
Overall, about 26\% of the Prolog solutions in the training set involves the $findall(.)$ predicate. Using $findall(.)$ departs from the chain of thought solution which is mainly verbal natural language reasoning without invoking constraint search.

\subsection{Solution Diversity}\label{diversity}
For mathematical reasoning problem, there may exists multiple solutions with variations. For example, a problem ``How many ways are there to put 5 balls in 2 boxes if the balls are not distinguishable but the boxes are?'', one Prolog solution employs $findall(.)$ and another solution uses $combination(.)$.
Our preliminary experiment shows that even we use a finetuned large language model to generate solutions on training questions, the generated solutions can be different from the original training references.
We believe that improving solution diversity will help improve the performance of a large language model especially for data augmentation.
Therefore, we devise a cross-validated self-training procedure to sample multiple different solutions during model finetuning. Algorithm~\ref{alg:cvst} shows the self-training procedure that requires a training and a generation set without overlap. For K-fold training, we set the k-th fold as a generation set and the rest as a training set.
\begin{algorithm}
  \SetAlgoLined
    \caption{A Self-Training Procedure.}
    \label{alg:cvst}
    \KwIn{Total training epochs $I$, Max solutions $J$, Initial model $m_0$.}
    \KwData {Train set $D$, Generation set $G$.} 
    \KwOut{Discovered set $S$.}
    \BlankLine
    \For{i in $I$}{
        $m_i$ $\leftarrow$ update($m_{i-1}$,$D$)\;
        \For{(q,a) in $G$}{
            $\hat{a} \leftarrow$ sample($m_i$,q)\;
            \If {$eval(\hat{a}) == eval(a)$ and $\hat{a} \ne a$}
            {Add (q,$\hat{a}$) into $S$ and $D$\;
            Remove (q,a) from $G$ when J different solutions are found.}
        }
        Shuffle $D$.
      }
\end{algorithm}

\section{Experiment}
\subsection{Setup}
\paragraph{Dataset}
We used the MATH corpus~\cite{hendrycksmath2021} and chose the counting and probability domain described in Section \ref{sec:Corpus} to construct the MATH-Prolog corpus. 
Initially, we employed GPT-4 to automatically generate Prolog codes with background operators for domain-specific questions, followed by manual correction to ensure accuracy. Prior to implementing the self-training procedure, we applied a filtering process to exclude training instances that could not be resolved using the provided numerical data or domain-specific commonsense knowledge. This resulted in a final set of 625 well-constructed Prolog samples, which we denote as $D$. To implement the self-training procedure described in \ref{diversity}, we partitioned the dataset into five folds, denoted $D_1, \dots, D_5$. The input prompting format follows the instruction prompts used in Stanford Alpaca \cite{alpaca}. 
We removed samples exceeding 2700 tokens to ensure that a training run can be fit into a 4090 GPU. All background operators were included in an input prompt to measure its effect. (see Appendix~\ref{appendix:ip} for details).

\paragraph{Training}
We conducted all experiments using the LLaMA-3.1 8B model \cite{dubey2024llama3herdmodels}. To reduce memory consumption while preserving performance, we employed 8-bit quantization and LoRA \cite{hu2022lora} for efficient model training.
We divided the training set into five folds and applied self-training to sample two correct samples per question on the heldout fold.
We experimented self-training using the test set as a generation set to discover Prolog solutions. See Appendix~\ref{sec:gpu} for further training details.

\paragraph{Evaluation}
We report accuracy (correct solution coverage) as a metric: 
\[\text{Acc} = \frac{\sum_{i=1}^{K}\sum_{j=1}^{|D_{i}|} \mathbbm{1}_{\left\{\mathcal{P}^{(a)}(s^{pred}_j)=\mathcal{P}^{(a)}(s^{true}_j)\right\}}}{|D|}\]
where $\mathcal{P}^{(a)}$ refers to the Swi-Prolog interpreter
, and $K$ is set to 5 folds. The metric measures the ability of a large language model to discover Prolog solutions in the unseen dataset.

\subsection{Results}
\begin{figure}[H]
  \centering
  \includegraphics[width=\columnwidth]{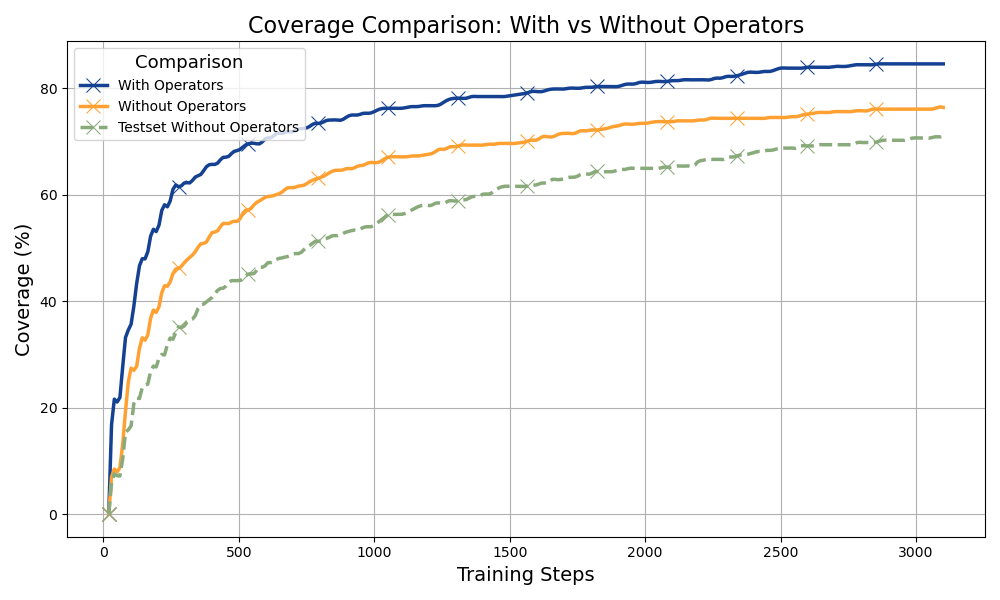}  
  \caption{Accuracy for 5-fold cross-validated self-training.}
  \label{figcross}
\end{figure}

\paragraph{Self-training is effective.} As shown in Figure \ref{figcross}, the coverage of correct generation achieves $84.6\%$ for the 5-fold cross-validation experiment. When we used the test set in self-training, test set accuracy was 75.7\% with background operators listed in the input context compared to 71.1\% without background operators after running for 3900 self-training steps. This showed that the introduction of background operators in the input context helped, and agreed with what were observed in the cross-validated self-training. When we used the derived test solutions from the first-round self-training and augmented them into the training set and rerun self-training again, the test set accuracy was increased to 84.8\%. We observed that some generated solutions exploits the compositional and decompositional nature of predicates (See \ref{sub:diversity} for examples). Their equivalence 
represent different problem-solving strategies giving diverse and high-quality solutions.


\paragraph{Including background operators in prompting helps.} As shown in Figure \ref{figcross}, using background operators in an input prompt exhibits a more efficient learning trajectory. By inspection, the model first tackled easy questions that involves one operator in early training steps, while complicated questions were tackled that involves multiple predicates at later training steps. For solution discovery sake, more training steps were preferable.


\section{Related Work}
\paragraph{Structured Reasoning}
The Chain-of-Thought (COT) prompting approach \cite{wei2022chain} was among the first attempts toward generating logical, verbal step-by-step reasoning. Several advanced techniques have been developed to enhance the reasoning capabilities \cite{zhou2023leasttomost, Zhu_2023, huang2022large, liang2023encouraging}. However, most of these approaches rely on natural language reasoning, which faces limitations in identifying flaws in verbal reasonings. To overcome this bottleneck, a notable attempt involves utilizing trees and graphs for reasoning, as graph-based structures \cite{zhang-etal-2020-graph-tree} have demonstrated potential in solving mathematical word problems~\cite{Kipf2016SemiSupervisedCW}. \cite{yu2023metamath,jiang-etal-2024-forward} employed forward and backward data augmentation via variable and final answer masking on COT solutions for data augmentation.
Our work extends the concept of graphical reasoning using predefined predicates to generate diverse and computable solutions.
\paragraph{Tool-based Reasoning}
External tools can be integrated into large language models to effectively enhance both the reasoning capabilities and interpretability of LLMs\cite{cobbe2021training, mishra2023lila, gou2023tora, gao2023pal, pmlr-v202-shao23a, chen2023program,AlphaGeometryTrinh2024}. Prolog, as a symbolic declarative language, maximizes the advantages for symbolic reasoning. It not only improves the logical coherence of natural language generation \cite{10.1007/978-3-031-71170-1_23} but also strengthens the model's ability in arithmetic reasoning \cite{yang-etal-2024-arithmetic,tan2024thoughtlikeproenhancingreasoninglarge,borazjanizadeh2024reliablereasoningnaturallanguage}. Our work employs standardized background operators for solving competition-level mathematical problems.

\section{Conclusion}
Our work introduces MATH-Prolog, a dataset specifically designed for tackling competition-level mathematical problems using declarative logic programming. With predefined predicates, we create a structured and interpretable framework for solving mathematical problems. Our cross-validated self-training algorithm effectively generates full solutions with computable intermediate steps on unseen questions. Our experiments demonstrates its efficiency and accuracy, showcasing its potential for broad application in systematic mathematical problem-solving.

\section{Limitation}
While our approach of incorporating predefined background operators provides a productive and interpretable method to define standardized solutions, its applicability has not yet been tested in other domains. Currently, we have only developed Prolog codes for the counting and probability domain, which may limit the scope of our experiments. However, we believe that the method can be generalized to  other domains within the MATH corpus. Additionally, we have observed that large language models occasionally generate Prolog code with syntax errors, which reduces the accuracy of code generation. Future work would involve expanding this approach to other domains of the MATH corpus. Moreover, constrained decoding~\cite{lu-etal-2022-neurologic,geng2023grammarconstrained} to eliminate syntax errors in generated code deserves further exploration. Lastly, our experiments have been limited to LLaMA-3.1 8B model. Future research would explore the effectiveness of our method to other models.







\bibliography{custom}

\begin{thebibliography}{30}
\providecommand{\natexlab}[1]{#1}

\bibitem[{Besta et~al.(2023)Besta, Blach, Kubicek, Gerstenberger, Gianinazzi,
  Gajda, Lehmann, Podstawski, Niewiadomski, Nyczyk, and
  Hoefler}]{besta2023graph}
Maciej Besta, Nils Blach, Ales Kubicek, Robert Gerstenberger, Lukas Gianinazzi,
  Joanna Gajda, Tomasz Lehmann, Michal Podstawski, Hubert Niewiadomski, Piotr
  Nyczyk, and Torsten Hoefler. 2023.
\newblock \href {https://arxiv.org/abs/2308.09687} {Graph of thoughts: Solving
  elaborate problems with large language models}.
\newblock \emph{Preprint}, arXiv:2308.09687.

\bibitem[{Borazjanizadeh and
  Piantadosi(2024)}]{borazjanizadeh2024reliablereasoningnaturallanguage}
Nasim Borazjanizadeh and Steven~T. Piantadosi. 2024.
\newblock \href {https://arxiv.org/abs/2407.11373} {Reliable reasoning beyond
  natural language}.
\newblock \emph{Preprint}, arXiv:2407.11373.

\bibitem[{Chen et~al.(2023)Chen, Ma, Wang, and Cohen}]{chen2023program}
Wenhu Chen, Xueguang Ma, Xinyi Wang, and William~W. Cohen. 2023.
\newblock \href {https://arxiv.org/abs/2211.12588} {Program of thoughts
  prompting: Disentangling computation from reasoning for numerical reasoning
  tasks}.
\newblock \emph{Preprint}, arXiv:2211.12588.

\bibitem[{Cobbe et~al.(2021)Cobbe, Kosaraju, Bavarian, Chen, Jun, Kaiser,
  Plappert, Tworek, Hilton, Nakano, Hesse, and Schulman}]{cobbe2021training}
Karl Cobbe, Vineet Kosaraju, Mohammad Bavarian, Mark Chen, Heewoo Jun, Lukasz
  Kaiser, Matthias Plappert, Jerry Tworek, Jacob Hilton, Reiichiro Nakano,
  Christopher Hesse, and John Schulman. 2021.
\newblock \href {https://arxiv.org/abs/2110.14168} {Training verifiers to solve
  math word problems}.
\newblock \emph{Preprint}, arXiv:2110.14168.

\bibitem[{et. al.(2024)}]{dubey2024llama3herdmodels}
Abhimanyu~Dubey et. al. 2024.
\newblock \href {https://arxiv.org/abs/2407.21783} {The llama 3 herd of
  models}.
\newblock \emph{Preprint}, arXiv:2407.21783.

\bibitem[{Gao et~al.(2023)Gao, Madaan, Zhou, Alon, Liu, Yang, Callan, and
  Neubig}]{gao2023pal}
Luyu Gao, Aman Madaan, Shuyan Zhou, Uri Alon, Pengfei Liu, Yiming Yang, Jamie
  Callan, and Graham Neubig. 2023.
\newblock \href {https://arxiv.org/abs/2211.10435} {Pal: Program-aided language
  models}.
\newblock \emph{Preprint}, arXiv:2211.10435.

\bibitem[{Geng et~al.(2023)Geng, Josifoski, Peyrard, and
  West}]{geng2023grammarconstrained}
Saibo Geng, Martin Josifoski, Maxime Peyrard, and Robert West. 2023.
\newblock \href {https://openreview.net/forum?id=KkHY1WGDII}
  {Grammar-constrained decoding for structured {NLP} tasks without finetuning}.
\newblock In \emph{The 2023 Conference on Empirical Methods in Natural Language
  Processing}.

\bibitem[{Gou et~al.(2023)Gou, Shao, Gong, Shen, Yang, Huang, Duan, and
  Chen}]{gou2023tora}
Zhibin Gou, Zhihong Shao, Yeyun Gong, Yelong Shen, Yujiu Yang, Minlie Huang,
  Nan Duan, and Weizhu Chen. 2023.
\newblock \href {https://arxiv.org/abs/2309.17452} {Tora: A tool-integrated
  reasoning agent for mathematical problem solving}.
\newblock \emph{Preprint}, arXiv:2309.17452.

\bibitem[{Hendrycks et~al.(2021)Hendrycks, Burns, Kadavath, Arora, Basart,
  Tang, Song, and Steinhardt}]{hendrycksmath2021}
Dan Hendrycks, Collin Burns, Saurav Kadavath, Akul Arora, Steven Basart, Eric
  Tang, Dawn Song, and Jacob Steinhardt. 2021.
\newblock Measuring mathematical problem solving with the math dataset.
\newblock \emph{NeurIPS}.

\bibitem[{Hu et~al.(2022)Hu, Shen, Wallis, Allen-Zhu, Li, Wang, Wang, and
  Chen}]{hu2022lora}
Edward~J Hu, Yelong Shen, Phillip Wallis, Zeyuan Allen-Zhu, Yuanzhi Li, Shean
  Wang, Lu~Wang, and Weizhu Chen. 2022.
\newblock \href {https://openreview.net/forum?id=nZeVKeeFYf9} {Lo{RA}: Low-rank
  adaptation of large language models}.
\newblock In \emph{International Conference on Learning Representations}.

\bibitem[{Huang et~al.(2022)Huang, Gu, Hou, Wu, Wang, Yu, and
  Han}]{huang2022large}
Jiaxin Huang, Shixiang~Shane Gu, Le~Hou, Yuexin Wu, Xuezhi Wang, Hongkun Yu,
  and Jiawei Han. 2022.
\newblock \href {https://arxiv.org/abs/2210.11610} {Large language models can
  self-improve}.
\newblock \emph{Preprint}, arXiv:2210.11610.

\bibitem[{Jiang et~al.(2024)Jiang, Shi, Yu, Liu, Zhang, Li, and
  Kwok}]{jiang-etal-2024-forward}
Weisen Jiang, Han Shi, Longhui Yu, Zhengying Liu, Yu~Zhang, Zhenguo Li, and
  James Kwok. 2024.
\newblock \href {https://aclanthology.org/2024.findings-acl.397}
  {Forward-backward reasoning in large language models for mathematical
  verification}.
\newblock In \emph{Findings of the Association for Computational Linguistics
  ACL 2024}, pages 6647--6661, Bangkok, Thailand and virtual meeting.
  Association for Computational Linguistics.

\bibitem[{Kipf and Welling(2016)}]{Kipf2016SemiSupervisedCW}
Thomas Kipf and Max Welling. 2016.
\newblock \href {https://api.semanticscholar.org/CorpusID:3144218}
  {Semi-supervised classification with graph convolutional networks}.
\newblock \emph{ArXiv}, abs/1609.02907.

\bibitem[{Liang et~al.(2023)Liang, He, Jiao, Wang, Wang, Wang, Yang, Tu, and
  Shi}]{liang2023encouraging}
Tian Liang, Zhiwei He, Wenxiang Jiao, Xing Wang, Yan Wang, Rui Wang, Yujiu
  Yang, Zhaopeng Tu, and Shuming Shi. 2023.
\newblock \href {https://arxiv.org/abs/2305.19118} {Encouraging divergent
  thinking in large language models through multi-agent debate}.
\newblock \emph{Preprint}, arXiv:2305.19118.

\bibitem[{Lu et~al.(2022)Lu, Welleck, West, Jiang, Kasai, Khashabi, Le~Bras,
  Qin, Yu, Zellers, Smith, and Choi}]{lu-etal-2022-neurologic}
Ximing Lu, Sean Welleck, Peter West, Liwei Jiang, Jungo Kasai, Daniel Khashabi,
  Ronan Le~Bras, Lianhui Qin, Youngjae Yu, Rowan Zellers, Noah~A. Smith, and
  Yejin Choi. 2022.
\newblock \href {https://doi.org/10.18653/v1/2022.naacl-main.57}
  {{N}euro{L}ogic a*esque decoding: Constrained text generation with lookahead
  heuristics}.
\newblock In \emph{Proceedings of the 2022 Conference of the North American
  Chapter of the Association for Computational Linguistics: Human Language
  Technologies}, pages 780--799, Seattle, United States. Association for
  Computational Linguistics.

\bibitem[{Mishra et~al.(2023)Mishra, Finlayson, Lu, Tang, Welleck, Baral,
  Rajpurohit, Tafjord, Sabharwal, Clark, and Kalyan}]{mishra2023lila}
Swaroop Mishra, Matthew Finlayson, Pan Lu, Leonard Tang, Sean Welleck, Chitta
  Baral, Tanmay Rajpurohit, Oyvind Tafjord, Ashish Sabharwal, Peter Clark, and
  Ashwin Kalyan. 2023.
\newblock \href {https://arxiv.org/abs/2210.17517} {Lila: A unified benchmark
  for mathematical reasoning}.
\newblock \emph{Preprint}, arXiv:2210.17517.

\bibitem[{Shao et~al.(2023)Shao, Gong, Shen, Huang, Duan, and
  Chen}]{pmlr-v202-shao23a}
Zhihong Shao, Yeyun Gong, Yelong Shen, Minlie Huang, Nan Duan, and Weizhu Chen.
  2023.
\newblock \href {https://proceedings.mlr.press/v202/shao23a.html} {Synthetic
  prompting: Generating chain-of-thought demonstrations for large language
  models}.
\newblock In \emph{Proceedings of the 40th International Conference on Machine
  Learning}, volume 202 of \emph{Proceedings of Machine Learning Research},
  pages 30706--30775. PMLR.

\bibitem[{Tan et~al.(2024)Tan, Deng, Qiu, Xu, Qu, Chu, Xu, and
  Qi}]{tan2024thoughtlikeproenhancingreasoninglarge}
Xiaoyu Tan, Yongxin Deng, Xihe Qiu, Weidi Xu, Chao Qu, Wei Chu, Yinghui Xu, and
  Yuan Qi. 2024.
\newblock \href {https://arxiv.org/abs/2407.14562} {Thought-like-pro: Enhancing
  reasoning of large language models through self-driven prolog-based
  chain-of-thought}.
\newblock \emph{Preprint}, arXiv:2407.14562.

\bibitem[{Taori et~al.(2023)Taori, Gulrajani, Zhang, Dubois, Li, Guestrin,
  Liang, and Hashimoto}]{alpaca}
Rohan Taori, Ishaan Gulrajani, Tianyi Zhang, Yann Dubois, Xuechen Li, Carlos
  Guestrin, Percy Liang, and Tatsunori~B. Hashimoto. 2023.
\newblock Stanford alpaca: An instruction-following llama model.
\newblock \url{https://github.com/tatsu-lab/stanford_alpaca}.

\bibitem[{Trinh et~al.(2024)Trinh, Wu, Le, He, and
  Luong}]{AlphaGeometryTrinh2024}
Trieu Trinh, Yuhuai Wu, Quoc Le, He~He, and Thang Luong. 2024.
\newblock \href {https://doi.org/10.1038/s41586-023-06747-5} {Solving olympiad
  geometry without human demonstrations}.
\newblock \emph{Nature}.

\bibitem[{Vakharia et~al.(2024)Vakharia, Kufeldt, Meyers, Lane, and
  Gilpin}]{10.1007/978-3-031-71170-1_23}
Priyesh Vakharia, Abigail Kufeldt, Max Meyers, Ian Lane, and Leilani~H. Gilpin.
  2024.
\newblock Proslm: A prolog synergized language model for explainable domain
  specific knowledge based question answering.
\newblock In \emph{Neural-Symbolic Learning and Reasoning}, pages 291--304,
  Cham. Springer Nature Switzerland.

\bibitem[{Wang et~al.(2024)Wang, Li, and Lu}]{wang2024dpost}
Tianduo Wang, Shichen Li, and Wei Lu. 2024.
\newblock Self-training with direct preference optimization improves
  chain-of-thought reasoning.
\newblock In \emph{Proceedings of ACL}.

\bibitem[{Wei et~al.(2022)Wei, Wang, Schuurmans, Bosma, Xia, Chi, Le, Zhou
  et~al.}]{wei2022chain}
Jason Wei, Xuezhi Wang, Dale Schuurmans, Maarten Bosma, Fei Xia, Ed~Chi, Quoc~V
  Le, Denny Zhou, et~al. 2022.
\newblock Chain-of-thought prompting elicits reasoning in large language
  models.
\newblock \emph{Advances in Neural Information Processing Systems},
  35:24824--24837.

\bibitem[{Yang et~al.(2024)Yang, Chen, and Tam}]{yang-etal-2024-arithmetic}
Xiaocheng Yang, Bingsen Chen, and Yik-Cheung Tam. 2024.
\newblock \href {https://doi.org/10.18653/v1/2024.naacl-short.61} {Arithmetic
  reasoning with {LLM}: {P}rolog generation {\&} permutation}.
\newblock In \emph{Proceedings of the 2024 Conference of the North American
  Chapter of the Association for Computational Linguistics: Human Language
  Technologies (Volume 2: Short Papers)}, pages 699--710, Mexico City, Mexico.
  Association for Computational Linguistics.

\bibitem[{Yao et~al.(2023)Yao, Yu, Zhao, Shafran, Griffiths, Cao, and
  Narasimhan}]{yao2023tree}
Shunyu Yao, Dian Yu, Jeffrey Zhao, Izhak Shafran, Thomas~L. Griffiths, Yuan
  Cao, and Karthik Narasimhan. 2023.
\newblock \href {https://arxiv.org/abs/2305.10601} {Tree of thoughts:
  Deliberate problem solving with large language models}.
\newblock \emph{Preprint}, arXiv:2305.10601.

\bibitem[{Yu et~al.(2023)Yu, Jiang, Shi, Yu, Liu, Zhang, Kwok, Li, Weller, and
  Liu}]{yu2023metamath}
Longhui Yu, Weisen Jiang, Han Shi, Jincheng Yu, Zhengying Liu, Yu~Zhang,
  James~T Kwok, Zhenguo Li, Adrian Weller, and Weiyang Liu. 2023.
\newblock Metamath: Bootstrap your own mathematical questions for large
  language models.
\newblock In \emph{International Conference on Learning Representations}.

\bibitem[{Zhang et~al.(2020)Zhang, Wang, Lee, Bin, Wang, Shao, and
  Lim}]{zhang-etal-2020-graph-tree}
Jipeng Zhang, Lei Wang, Roy Ka-Wei Lee, Yi~Bin, Yan Wang, Jie Shao, and Ee-Peng
  Lim. 2020.
\newblock \href {https://doi.org/10.18653/v1/2020.acl-main.362} {Graph-to-tree
  learning for solving math word problems}.
\newblock In \emph{Proceedings of the 58th Annual Meeting of the Association
  for Computational Linguistics}, pages 3928--3937, Online. Association for
  Computational Linguistics.

\bibitem[{Zhihong~Shao(2024)}]{deepseek-math}
Qihao Zhu Runxin Xu Junxiao Song Mingchuan Zhang Y.K. Li Y. Wu Daya~Guo
  Zhihong~Shao, Peiyi~Wang. 2024.
\newblock \href {https://arxiv.org/abs/2402.03300} {Deepseekmath: Pushing the
  limits of mathematical reasoning in open language models}.
\newblock \emph{CoRR}, abs/2402.03300.

\bibitem[{Zhou et~al.(2023)Zhou, Schärli, Hou, Wei, Scales, Wang, Schuurmans,
  Cui, Bousquet, Le, and Chi}]{zhou2023leasttomost}
Denny Zhou, Nathanael Schärli, Le~Hou, Jason Wei, Nathan Scales, Xuezhi Wang,
  Dale Schuurmans, Claire Cui, Olivier Bousquet, Quoc Le, and Ed~Chi. 2023.
\newblock \href {https://arxiv.org/abs/2205.10625} {Least-to-most prompting
  enables complex reasoning in large language models}.
\newblock \emph{Preprint}, arXiv:2205.10625.

\bibitem[{Zhu et~al.(2023)Zhu, Wang, Zhang, Zhang, Huang, Gan, Zhang, and
  Yang}]{Zhu_2023}
Xinyu Zhu, Junjie Wang, Lin Zhang, Yuxiang Zhang, Yongfeng Huang, Ruyi Gan,
  Jiaxing Zhang, and Yujiu Yang. 2023.
\newblock \href {https://doi.org/10.18653/v1/2023.acl-long.245} {Solving math
  word problems via cooperative reasoning induced language models}.
\newblock In \emph{Proceedings of the 61st Annual Meeting of the Association
  for Computational Linguistics (Volume 1: Long Papers)}. Association for
  Computational Linguistics.

\end{thebibliography}

\newpage
\onecolumn
\appendix

\section{Appendix}
\label{sec:appendix}

\subsection{Diversity Examples}
\label{sub:diversity}
\begin{figure*}[htbp]
  \centering
  \includegraphics[width=1.08\linewidth]{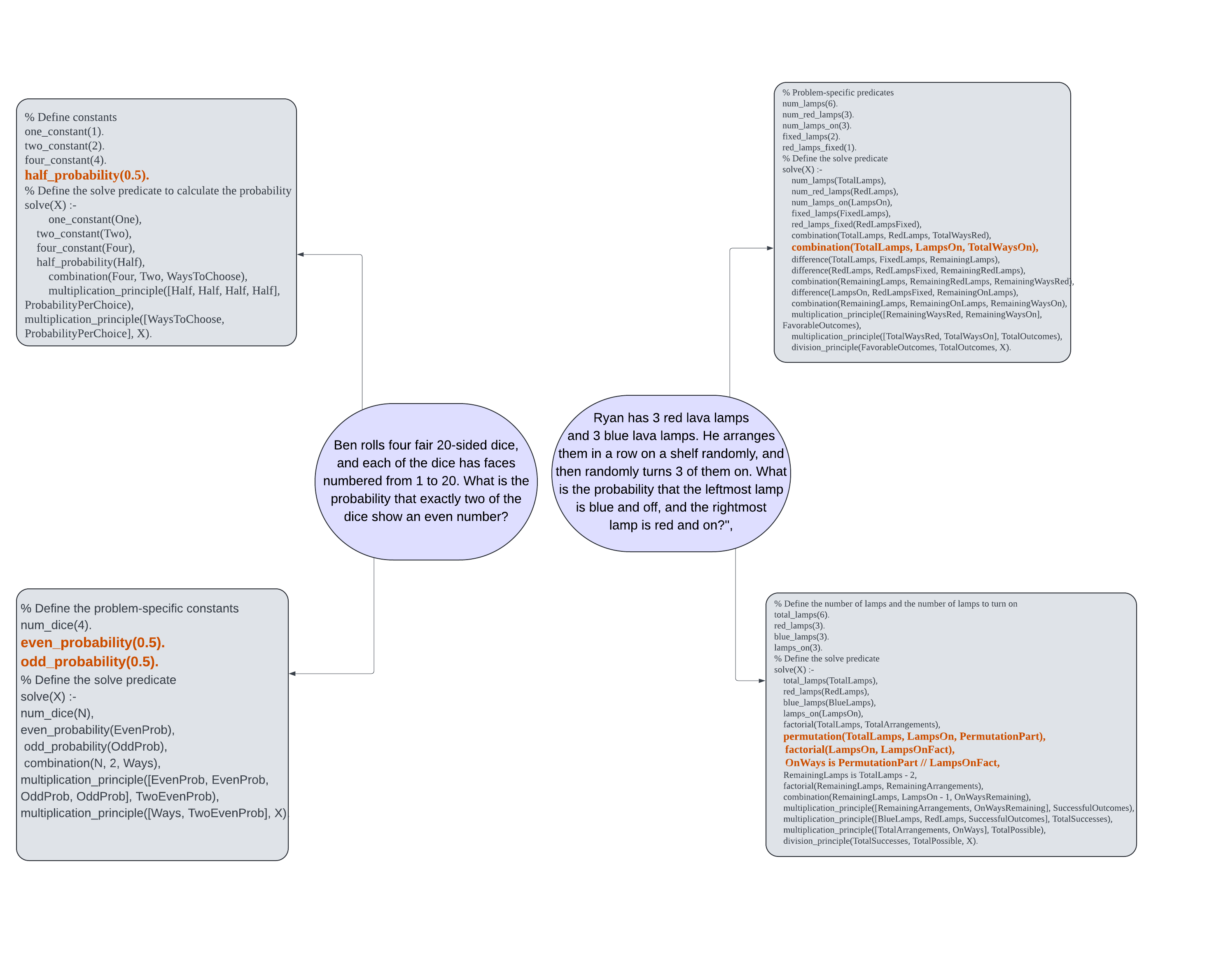} 
  \caption{Diversity between original solution and augmented solution.}
  \label{Diversity}
\end{figure*}


In Figure \ref{Diversity}, consider the question: "Ben rolls four fair 20-sided dice, each with faces numbered from 1 to 20. What is the probability that exactly two of the dice show an even number?" Initially, we define the predicate $half\_probability(0.5)$ to represent the likelihood of a specific outcome for a die. However, after applying the cross-validated self-training procedure, the model identifies an alternative interpretation of the problem, where the probability of observing either an even or odd number is both $0.5$. This result highlights the model's ability to extract implicit information from the problem statement, which can deviate from our original understanding.

When generating a solution for a question such as "Ryan has 3 red lava lamps and 3 blue lava lamps. He arranges them randomly in a row on a shelf, and then randomly turns 3 of them on. What is the probability that the leftmost lamp is blue and off, and the rightmost lamp is red and on?", the model discovers an alternative solution strategy. Graphically, the new solution involves a decomposition of the combination operator. If we denote the combination operator as $Com$, the permutation operator as $Per$, the factorial operator as $Fac$, and the division operator as $Div$, then $Com = Per \oplus Fac \oplus Div$. This type of decomposition offers a structured and robust approach for augmenting new diverse samples.

\subsection{background operators}
Below, we provide a detailed overview of the predicates used in the counting and probability domain of the MATH-Prolog corpus. We describe the input-output behavior of each operator, as well as its potential for decomposition, which is leveraged in the data augmentation process.

\begin{longtable}{p{10cm}|p{6cm}}
    \toprule
    \textbf{Operator/Predicate} & \textbf{Decomposition of Operator} \\ 
    \midrule
    \endfirsthead
    \toprule
    \textbf{Operator/Predicate} & \textbf{Decomposition of Operator} \\ 
    \midrule
    \endhead
    \midrule
    \endfoot
    \bottomrule
    \endlastfoot
    \texttt{max(A, B, Max)}: Find the maximum of two numbers \newline 
    Inputs: A (number), B (number) \newline Output: Max (number) & None \\ \hline

    \texttt{member(X, List)}: Check if an element is in a list \newline 
    Inputs: X (element), List (list) \newline Output: true/false & None \\ \hline

    \texttt{subset(A, B)}: Check if list A is a subset of list B \newline 
    Inputs: A (list), B (list) \newline Output: true/false &  None\\ \hline

    \texttt{sets\_equal(A, B)}: Check if two sets are equal \newline 
    Inputs: A (list), B (list) \newline Output: true/false &  $sets\_equal = member \oplus subset$\\ \hline

    \texttt{proper\_subset(A, B)}: Check if list A is a proper subset of list B \newline 
    Inputs: A (list), B (list) \newline Output: true/false & $proper\_subset = sets\_equal \oplus sets\_equal$ \\ \hline

    \texttt{union(A, B, U)}: Get the union of two sets \newline 
    Inputs: A (list), B (list) \newline Output: U (list) &  implemented by member\\ \hline

    \texttt{intersection(A, B, I)}: Get the intersection of two sets \newline 
    Inputs: A (list), B (list) \newline Output: I (list) &  implemented by member\\ \hline

    \texttt{complement(A, U, C)}: Get the complement of a set A with respect to a universal set U \newline 
    Inputs: A (list), U (list) \newline Output: C (list) &  implemented by member\\ \hline

    \texttt{cardinality(Set, N)}: Count the number of elements in a set or measure the size of an area \newline 
    Inputs: Set (list or number) \newline Output: N (number) &  None\\ \hline

    \texttt{card\_union(A, B, N)}: Calculate the cardinality of the union of two sets \newline 
    Inputs: A (list), B (list) \newline Output: N (number) &  $card\_union = cardinality \oplus union$\\ \hline

    \texttt{card\_intersection(A, B, N)}: Calculate the cardinality of the intersection of two sets \newline 
    Inputs: A (list), B (list) \newline Output: N (number) &  $card\_intersection = intersection \oplus cardinality$\\ \hline

    \texttt{classical\_probability\_model(SampleSpace)}: Check if the sample space is a classical probability model \newline 
    Inputs: SampleSpace (list) \newline Output: true/false &  implemented by cardinality\\ \hline

    \texttt{probability\_of\_event(Event, SampleSpace, P)}: Calculate the probability of an event \newline 
    Inputs: Event (list), SampleSpace (list) \newline Output: P (number) &  $probability\_of\_event = cardinality\oplus classical\_probability\_model$\\ \hline

    \texttt{mutually\_exclusive(EventA, EventB)}: Check if two events are mutually exclusive \newline 
    Inputs: EventA (list), EventB (list) \newline Output: true/false &  implemented by intersection\\ \hline

    \texttt{probability\_of\_union(EventA, EventB, SampleSpace, P)}: Calculate the probability of the union of two events \newline 
    Inputs: EventA (list), EventB (list), SampleSpace (list) \newline Output: P (number) & $probability\_of\_union = probability\_of\_event \oplus intersection$ \\ \hline

    \texttt{independent\_events(EventA, EventB, SampleSpace)}: Check if two events are independent \newline 
    Inputs: EventA (list), EventB (list), SampleSpace (list) \newline Output: true/false & $independent\_events = probability\_of\_event \oplus intersection$  \\ \hline

    \texttt{division\_principle(A, B, Result)}: Apply the division principle \newline 
    Inputs: A (number or list), B (number or list) \newline Output: Result (number or list) &  None\\ \hline

     \texttt{difference(A, B, Diff)}: Calculate the difference between two numbers or two lists \newline 
    Inputs: A (number or list), B (number or list) \newline Output: Diff (number or list) &  None\\ \hline

    \texttt{factorial(N, F)}: Calculate the factorial of a number \newline 
    Inputs: N (number) \newline Output: F (number) &  None\\ \hline

    \texttt{permutation(N, R, P)}: Calculate permutations \newline 
    Inputs: N (number), R (number) \newline Output: P (number) &  $permutation = factorial \oplus division\_principle \oplus difference$\\ \hline

    \texttt{combination(N, R, C)}: Calculate combinations \newline 
    Inputs: N (number), R (number) \newline Output: C (number) &  $combination = permutation \oplus factorial \oplus division_principle$\\ \hline

    \texttt{multiplication\_principle(List, P)}: Apply the multiplication principle \newline 
    Inputs: List (list) \newline Output: P (number) &  None\\ \hline

    \texttt{addition\_principle(List, S)}: Apply the addition principle \newline 
    Inputs: List (list) \newline Output: S (number) &  None\\ \hline

    \texttt{power(Base, Exponent, Result)}: Calculate the power of a base to an exponent \newline 
    Inputs: Base (number), Exponent (number) \newline Output: Result (number) &  None \\ \hline

    \texttt{sumlist(List, Sum)}: Calculate the sum of a list of numbers \newline 
    Inputs: List (list) \newline Output: Sum (number) &  None\\ \hline

    \texttt{power\_set\_cardinality(Set, N)}: Calculate the number of elements in the power set \newline 
    Inputs: Set (list) \newline Output: N (number) &  $power\_set\_cardinality = cardinality \oplus power$\\ \hline

    \texttt{binomial\_expansion(A, B, N, Sum)}: Calculate the binomial expansion \newline 
    Inputs: A (number), B (number), N (number) \newline Output: Sum (number) &  $binomial\_expansion = findall(.) \oplus combination \oplus power \oplus sumlist$\\ \hline

    \texttt{prime\_factors(N, Factors)}: Perform prime factorization \newline 
    Inputs: N (number) \newline Output: Factors (list) &  implemented by $findall(.)$\\ \hline

    \texttt{pascal\_rule(N, K, Result)}: Calculate binomial coefficients using Pascal's Rule \newline 
    Inputs: N (number), K (number) \newline Output: Result (number) &  implemented by combination\\ \hline

    \texttt{area\_of\_rectangle(Width, Height, Area)}: Calculate the area of a rectangle \newline 
    Inputs: Width (number), Height (number) \newline Output: Area (number) & None \\ \hline

    \texttt{area\_of\_triangle(Base, Height, Area)}: Calculate the area of a triangle \newline 
    Inputs: Base (number), Height (number) \newline Output: Area (number) &  None\\ \hline

    \texttt{constant\_to\_list(Constant, List)}: Convert a constant to a list \newline 
    Inputs: Constant (element) \newline Output: List (list) &  None\\ \hline

    \texttt{last(List, X)}: Find the last element of a list \newline 
    Inputs: List (list) \newline Output: X (element) &  None\\ \hline

    \texttt{extract\_single\_value(List, Value)}: Extract a single value from a list \newline 
    Inputs: List (list) \newline Output: Value (element) &  implemented by last\\ \hline

    \texttt{probability\_complement(P, CompP)}: Calculate the complement probability \newline 
    Inputs: P (number) \newline Output: CompP (number) &  implemented by difference\\ \hline

    \texttt{solve\_linear\_equation(A, B, C, X)}: Solve the linear equation Ax + B = C \newline 
    Inputs: A (number), B (number), C (number) \newline Output: X (number) & $solve\_linear\_equation = difference \oplus division\_principle$ \\ \hline

    \texttt{can\_form\_triangle(Sides)}: Check if three stick lengths can form a triangle \newline 
    Inputs: Sides (list) \newline Output: true/false &  None\\ \hline

    \texttt{gcd(A, B, G)}: Compute the greatest common divisor \newline 
    Inputs: A (number), B (number) \newline Output: G (number) &  None\\ \hline

    \texttt{coprime(A, B)}: Check if two numbers are coprime \newline 
    Inputs: A (number), B (number) \newline Output: true/false &  implemented by gcd\\ \hline

    \texttt{number\_of\_diagonals(N, Diagonals)}: Calculate the number of diagonals in a polygon \newline 
    Inputs: N (number) \newline Output: Diagonals (number) &  implemented by division\_principle\\ \hline

    \texttt{expected\_value(Outcomes, Probability, ExpectedValue)}: Calculate the expected value \newline 
    Inputs: Outcomes (list), Probability (number or list) \newline Output: ExpectedValue (number) &  $expected\_value = findall(.) \oplus multiplication\_principle \oplus addition\_principle$\\ \hline

    \texttt{is\_prime(P)}: Check if a number is prime \newline 
    Inputs: P (number) \newline Output: true/false &  None\\ \hline

    \texttt{number\_digits(Number, Digits)}: Convert a number to its digits \newline 
    Inputs: Number (number) \newline Output: Digits (list) &  None\\ \hline

    \texttt{ceiling(X, CeilX)}: Calculate the ceiling of a number \newline 
    Inputs: X (number) \newline Output: CeilX (number) &  None\\ \hline

    \texttt{floor(X, FloorX)}: Calculate the floor of a number \newline 
    Inputs: X (number) \newline Output: FloorX (number) &  None\\ \hline

    \texttt{sqrt(X, Y)}: Calculate the square root of a number \newline 
    Inputs: X (number) \newline Output: Y (number) &  None\\ \hline

    \texttt{pi\_constant(Pi)}: Define the value of Pi None\newline 
    Output: Pi (number) &  \\ \hline

    \texttt{sphere\_volume(Radius, Volume)}: Calculate the volume of a sphere \newline 
    Inputs: Radius (number) \newline Output: Volume (number) &  $sphere\_volume = pi\_constant \oplus division\_principle \oplus power \oplus multiplication\_principle $\\ \hline

    \texttt{cube\_volume(Side, Volume)}: Calculate the volume of a cube \newline 
    Inputs: Side (number) \newline Output: Volume (number) &  implemented by power\\ \hline

    \texttt{mod(X, Y, Z)}: Calculate the remainder of division \newline 
    Inputs: X (number), Y (number) \newline Output: Z (number) &  None\\ \hline

    \texttt{palindrome(Number)}: Check if a number is a palindrome \newline 
    Inputs: Number (number) \newline Output: true/false &  None\\ \hline

    \texttt{fibonacci(N, Result)}: Calculate the Nth Fibonacci number using recursion \newline 
    Inputs: N (number) \newline Output: Result (number) &  None\\ \hline
\end{longtable}

\subsection{Instruction Prompt}
\label{appendix:ip}
Below are the instruction prompts we use for different settings:
\begin{table}[H]
    \centering
    \footnotesize
    \begin{tabular}{p{4cm}|p{11cm}}
    \toprule
    Setting & Prompt Template\\ \midrule
    Prolog generation without predicates & Below is an instruction that describes a task, paired with an input that provides further context. Write a response that appropriately completes the request.\\
                      & \#\#\# Instruction:\\
                      & Please generate a piece of Prolog code to solve the given math problem.\\
                      & \#\#\# Input:\\
                      & <Question>\\
                      & \#\#\# Output:\\
                      & <Prolog Code>\\ \midrule
    Prolog generation with predicates & Below is an instruction that describes a task, paired with an input that provides further context. Write a response that appropriately completes the request.\\
                               & \#\#\# Instruction:\\
                               & Please generate a prolog answer based on the given predicates to solve the given math problem.\\
                               & <background operators> \\
                               & \#\#\# Input:\\
                               & <Question>\\
                               & \#\#\# Output:\\
                               & <Prolog Code>\\
    \bottomrule
    \end{tabular}
\end{table}
\subsection{Training Samples}
\begin{table}[H]
    \centering
    \scriptsize
    \begin{tabular}[t]{p{4cm}|p{11cm}}
    \toprule
    \textbf{Question} & \textbf{Prolog Code}\\ \midrule

    Robert likes chocolate milk, so he decides to visit the milk bottling plant every day for a week to get the free samples. Unfortunately for him, the bottling plant sometimes bottles regular milk instead of chocolate milk, so each day the plant has a 2/3 chance of bottling chocolate milk. What is the probability that the bottling plant bottles chocolate milk exactly 4 of the 5 days he visits? & \begin{tabular}[c]{@{}l@{}} total\_days(5).\\ successful\_days(4).\\ failure\_days(1).\\ two\_thirds(2/3).\\ \\ solve(X) :-\\ \ \ \ total\_days(TotalDays),\\ \ \ \ successful\_days(SuccessfulDays),\\ \ \ \ failure\_days(FailureDays),\\ \ \ \ two\_thirds(TwoThirds),\\ \ \ \ combination(TotalDays, SuccessfulDays, NumWays),\\ \ \ \ power(TwoThirds, SuccessfulDays, SuccessProb),\\ \ \ \ probability\_complement(TwoThirds, OneThird),\\ \ \ \ power(OneThird, FailureDays, FailProb),\\ \ \ \ multiplication\_principle([NumWays, SuccessProb, FailProb], X).\end{tabular} \\ \midrule

    Juan rolls a fair regular octahedral die marked with the numbers 1 through 8. Then Amal rolls a fair six-sided die. What is the probability that the product of the two rolls is a multiple of 3? & \begin{tabular}[c]{@{}l@{}} one\_constant(1).\\ eight\_sides(8).\\ six\_sides(6).\\ modulus(3).\\ \\ solve(X) :-\\ \ \ \ one\_constant(One),\\ \ \ \ eight\_sides(Eight),\\ \ \ \ six\_sides(Six),\\ \ \ \ modulus(Modulus),\\ \ \ \ findall((J, A), (between(One, Eight, J), between(One, Six, A)), AllPairs),\\ \ \ \ findall(1, (member((J, A), AllPairs), Product is J * A, Product mod Modulus =:= 0), ValidPairs),\\ \ \ \ length(AllPairs, TotalPairs),\\ \ \ \ length(ValidPairs, ValidCount),\\ \ \ \ division\_principle(ValidCount, TotalPairs, X).\end{tabular} \\ \midrule

    Mr. Reader has six different Spiderman comic books, five different Archie comic books and four different Garfield comic books. When stacked, all of the Spiderman comic books are grouped together, all of the Archie comic books are grouped together and all of the Garfield comic books are grouped together. In how many different orders can these 15 comic books be stacked in a pile with the covers facing up and all of them facing the same direction? Express your answer as a whole number. & \begin{tabular}[c]{@{}l@{}} spiderman\_books(6).\\ archie\_books(5).\\ garfield\_books(4).\\ groups(3).\\ \\ solve(X) :-\\ \ \ \ spiderman\_books(SpidermanBooks),\\ \ \ \ archie\_books(ArchieBooks),\\ \ \ \ garfield\_books(GarfieldBooks),\\ \ \ \ groups(Groups),\\ \ \ \ factorial(SpidermanBooks, SpidermanWays),\\ \ \ \ factorial(ArchieBooks, ArchieWays),\\ \ \ \ factorial(GarfieldBooks, GarfieldWays),\\ \ \ \ factorial(Groups, GroupWays),\\ \ \ \ multiplication\_principle([SpidermanWays, ArchieWays, GarfieldWays, GroupWays], X).\end{tabular} \\

    \bottomrule
    \end{tabular}
\end{table}

\subsection{Error Analysis}
Below, we present some common errors generated by the LLaMA-3.1 8B model during the self-training procedure, to better understand the limitations and challenges in generating Prolog programs. These examples were selected from experiments that included predicates in the input prompts. The majority of problematic solutions exhibit syntax errors, while others fail to correctly identify and apply the constraints specified in the problem statements. A few typical syntax errors are listed below, with the erroneous lines highlighted in bold. Detailed explanations of each error are provided in the accompanying comments.

\begin{table}[H]
    \centering
    \scriptsize
    \begin{tabular}[t]{p{4cm}|p{11cm}}
    \toprule
    \textbf{Question} & \textbf{Prolog Code}\\ \midrule

    $n$ fair 6-sided dice are simultaneously rolled. The probability that exactly two of them show a number other than 1 is $\frac{25}{216}$. Find $n$. & \begin{tabular}[c]{@{}l@{}} 

solve(N) :-\\
    between(2, 10, N),\\
    findall(K, (\\
        between(0, N, K),\\
        \textbf{difference(N, K, L),} \\ 
        combination(N, K, CombK),\\
        power(5, K, Factor1),\\
        \textbf{power(1, L, Factor2),} \\
        multiplication\_principle([CombK, Factor1, Factor2], Temp),\\
        division$\_$principle(Temp, 6\^{}N, ProbK)$\%$ **Error: count compute power inside division here. **\\
    ), ProbList),\\
    findall(ProbN, (\\
        member(K, ProbList),\\
        combination(N, K, CombK),\\
        power(5, K, Factor1),\\
        \textbf{power(1, L, Factor2),} $\%$ **Error: L not correctly passed into the next predicates.**\\  \\
        multiplication\_principle([CombK, Factor1, Factor2], Temp),\\
        division\_principle(Temp, 6\^{}N, ProbN)\\
    ), ProbNList),\\
    findall(N, (\\
        member(ProbN, ProbNList),\\
        ProbN =:= 25/216\\
    ), [N]).\end{tabular} \\ \midrule

     The letters $A, B$ and $C$ are used to form every possible three letter ``word.'' When these ``words'' are arranged in alphabetical order and numbered so that $AAA$ is Word 1 and $CCC$ is Word 27, what number will correspond to the position of word $BAB$ on the list? & \begin{tabular}[c]{@{}l@{}} 

one(1).\\
two(2).\\
total\_words(27).\\

solve(X) :-\\
    one(One),\\
    two(Two),\\
    total\_words(TotalWords),\\
    \textbf{multiplication\_principle([TotalWords, One], Position),} $\%$ **Error: Not explicitly calculate for case $BAB$. **\\
    addition\_principle([Position, One], X).\\\end{tabular} \\ \midrule

    The Smith family has 4 sons and 3 daughters. In how many ways can they be seated in a row of 7 chairs such that all 3 girls sit next to each other? & \begin{tabular}[c]{@{}l@{}} 

total\_children(7).\\
girls(3).\\
empty\_slot(1).\\

solve(X) :-\\
    total\_children(TotalChildren),\\
    girls(Girls),\\
    empty\_slot(EmptySlot),\\
    \textbf{factorial(TotalChildren, TotalArrangements),} \\ 
    $\%$ **Error: This incorrectly calculates the total arrangement of all 7 children. \\
     The 3 girls should be treated as a single block, reducing the problem to arranging 5 elements (4 boys + 1 block of girls).** \\
    
    factorial(Girls, GirlArrangements),\\
    
    difference(TotalChildren, Girls, RemainingChildren),\\
    factorial(RemainingChildren, RemainingArrangements),\\
    \textbf{multiplication\_principle([TotalArrangements, GirlArrangements, RemainingArrangements], X).} \\
    $\%$**Error: This incorrectly includes the arrangement of all remaining children, \\
     whereas it should first calculate the arrangement of 5 elements, then multiply by the arrangement of 3 girls.**\\
\end{tabular} \\

    \bottomrule
    \end{tabular}
\end{table}

\subsection{Training Details and Computational Budget} \label{sec:gpu}
We applied LoRA to fine-tune the query and value weight matrices within the transformer blocks. After exploring various hyperparameter configurations, we selected a LoRA rank and scaling factor of $(r, \alpha) = (8, 16)$ to achieve an optimal trade-off between performance and computational efficiency. To collect diverse Prolog solutions, we ran the self-training experiments for 100 epochs with batch size of 16, and learning rate of $3\times10^{-4}$. For a self-training run, we used 1 NVIDIA RTX 4090 GPU to finetune Meta-Llama-3.1-8B-Instruct model on each fold, taking around 2 days to finish. The sampling hyperparameters include temperature scaling of $0.6$, top-k sampling of $40$, and nucleus sampling of $0.9$.



\end{document}